\documentclass[conference]{IEEEtran}

\usepackage{tikz,tkz-euclide,pgfplots}
\tikzstyle{roundnode} = [circle, fill=black!255, scale=.8]

\tikzstyle{squarenode} = [square, fill=black!255, scale=1.0]

\usepackage{pifont}
\usepackage{amsmath}
\usepackage{balance}
\usepackage[noadjust]{cite}
\usepackage{threeparttable}
\usepackage{xcolor}
\usepackage{soul}           
\usepackage[linesnumbered, ruled, vlined]{algorithm2e}
\let\oldnl\nl
\newcommand{\nonl}{\renewcommand{\nl}{\let\nl\oldnl}}
\usepackage{comment}

\ifCLASSINFOpdf
  \usepackage[]{graphicx}
\else
\fi
\usepackage{array}
\usepackage{algcompatible}

\usepackage{multirow}
\usepackage{graphicx}
\usepackage{color, soul}
\usepackage{color, colortbl}

\usepackage{makecell}

\usepackage{breqn}

\begin{document}
\bstctlcite{IEEEexample:BSTcontrol}
%
\title{Reliability-Aware Deployment of DNNs on In-Memory Analog Computing Architectures}

\author{\IEEEauthorblockN{Md Hasibul Amin, Mohammed Elbtity, Ramtin Zand}
\IEEEauthorblockA{Department of Computer Science and Engineering, University of South Carolina, Columbia, SC 29208, USA
}
}



%


\maketitle

\IEEEpeerreviewmaketitle

\section{Introduction}

Conventional in-memory computing (IMC) architectures consist of analog memristive crossbars to accelerate matrix-vector multiplication (MVM), and digital functional units to realize nonlinear vector (NLV) operations in deep neural networks (DNNs). These designs, however, require energy-hungry signal conversion units which can dissipate more than 95\% of the total power of the system. In-Memory Analog Computing (IMAC) circuits \cite{GLSVLSI_neuron, 9516756}, on the other hand, remove the need for signal converters by realizing both MVM and NLV operations in the analog domain leading to significant energy savings. However, they are more susceptible to reliability challenges such as interconnect parasitic and noise \cite{parasitic_iscas}. Here, we introduce a practical approach to deploy large matrices in DNNs onto multiple smaller IMAC subarrays to alleviate the impacts of noise and parasitics while keeping the computation in the analog domain.

\begin{figure}
\centering
\includegraphics[width=3.2in]{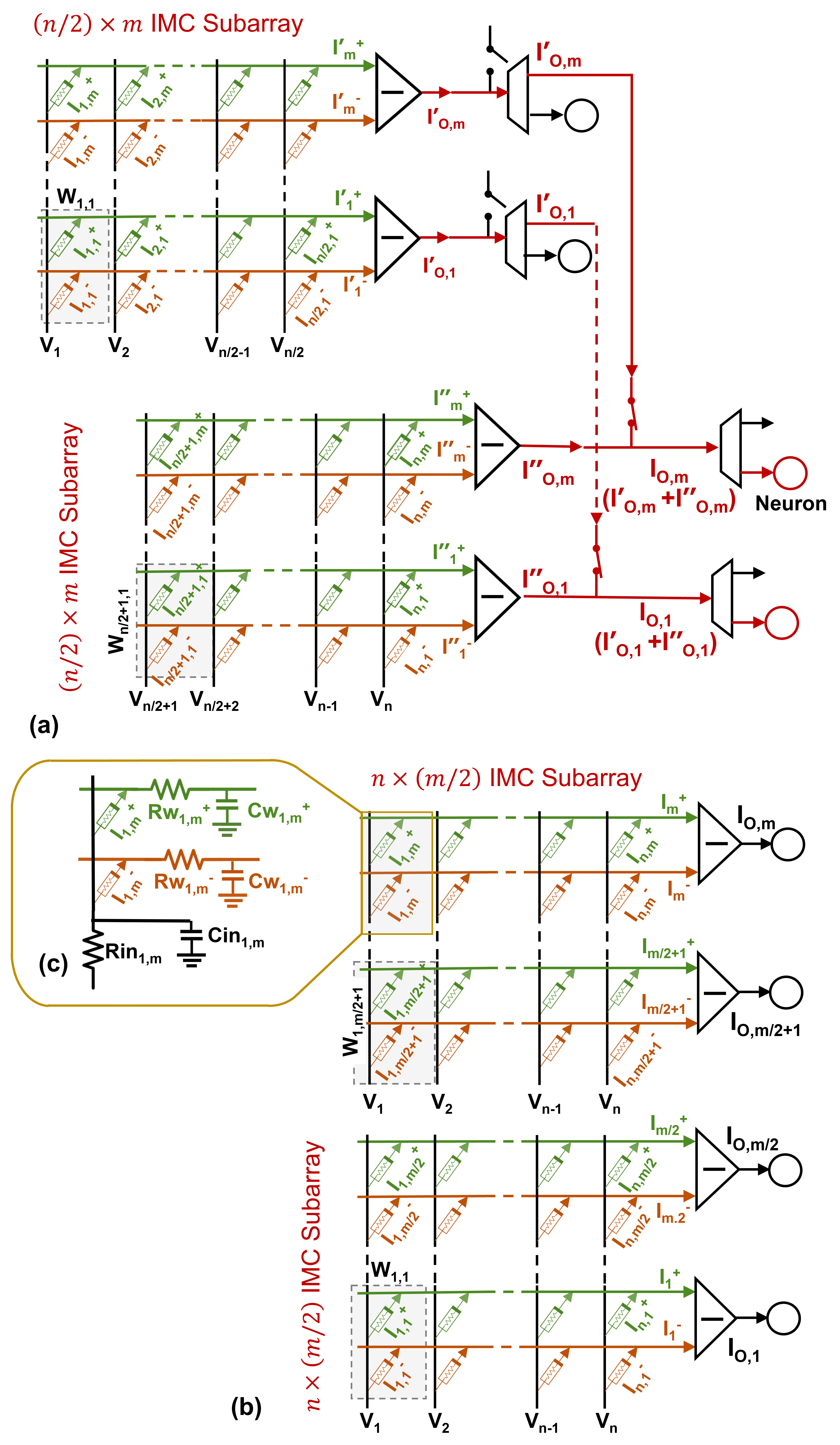}
\vspace{-1mm}
\caption{\small The Xbar-partitioning approach. (a) Horizontal partitioning ($H_P=2$), and (b) vertical partitioning ($V_P=2$) in an analog IMC array. (c) Parasitic capacitance and resistance model.}
\vspace{-6mm}
\label{fig:partition}
\end{figure}

\section{Proposed Partitioning Approach}

Here, we propose a partitioning method consisting of horizontal and vertical partitioning techniques including specialized routing circuitry to handle partitioning of IMAC circuits. Figures \ref{fig:partition} (a) and \ref{fig:partition} (b) provide a schematic of the horizontal and vertical partitioning circuitry, respectively. For the horizontal partitioning, a layer of demultiplexers (DEMUX) is added to the output of the crossbars, which distributes the output currents corresponding to the matrix-vector multiplication results to either neurons in the same subarray for normal non-partitioned operation, or to the next subarray as partial products of that particular partition. Moreover, we place switches on the output of the crossbars before DEMUX circuits to identify whether the generated output currents should be accumulated with the currents arriving from other subarrays (i.e. partitions) or not. Using these peripheral circuitry IMC circuit can handle the horizontal partitioning in the analog domain. Figure \ref{fig:partition} (a) shows an example of horizontal partitioning ($H_P$) of an $n\times m$ array into two $n/2\times m$ partitions. $I'_{o,i}$ and $I''_{o,i}$ currents are partial products that are obtained from first and second partitions, respectively. First, $I'_{o,i} = \sum^{n/2}_{k=1}  (G_{k,i}^+-G_{k,i}^-) V_k$ obtained from the first partition is passed to the second partition through the DEMUX and is accumulated with $I''_{o,i}=\sum^{n}_{k=n/2 +1}  (G_{k,i}^+-G_{k,i}^-) V_k$ in the second partition. Then, $I_{o,i}=I'_{o,i}+I''_{o,i}=\sum^{n}_{k=1}  (G_{k,i}^+-G_{k,i}^-) V_k$ is routed to the analog neuron as the result of accumulated MAC operations in both partitions. The vertical partitioning ($V_P$) is more straightforward than horizontal partitioning. Figure \ref{fig:partition} (b) shows a sample of vertical partitioning of $n\times m$ array into two $n\times m/2$ subarrays. Vertical partitioning only requires the output of all partitions to be concatenated in the switch blocks before transferring them to the next layer. In general, an $n \times m$ array can be vertically partitioned into multiple $n \times k_i$ subarrays, in which $m=\sum_{i=0}^{V_P} k_i$, where $V_P$ is the total number of vertical partitions. 

\section{Simulation Results and Discussion}
Here, we use the proposed partitioning mechanism to deploy a binarized $400\times120\times84\times10$ DNN model, pretrained for MNIST classification, onto various IMAC architectures with different subarray sizes, memristive technologies and bitcell types. We use SPICE circuit simulator and 14nm High-Performance PTM-MG FinFET technology to obtain the results exhibited in this section.

\vspace{-1mm}

\subsection{Partitioning for Parasitics Tolerance}

The bar graphs in Fig. \ref{fig:bar} (a) and \ref{fig:bar} (b) demonstrate the accuracy and power consumption results. The results show that with lower partitioning on $256\times256$ and $128\times128$ subarrays, the deployed model fails to provide reliable classification in all cases except for the PCM-based memristive crossbars \cite{in-memory-PCM}. For instance, the deployed DNN model on the IMC architecture constructed with $256\times256$ subarrays of 1T-1R STT-MRAM bitcells can barely obtain 11\% accuracy, while it can achieve 91.6\% accuracy when using $32\times32$ subarrays. The improvement in accuracy is an outcome of the increased number of horizontal and vertical partitions, as decreasing the size of the subarrays reduces the length of the interconnects and consequently their parasitic resistances. However, higher partitioning causes a high power consumption of 3.18W for the DNN deployment on $32\times32$ subarrays of 1T-1R STT-MRAM bitcells, while similar design with $256\times256$ subarrays only dissipates 0.996W of power. Hence, the high accuracy is achieved at the cost of increased power consumption due to the extra circuitry added to handle partitioning. 
\vspace{-1mm}
\subsection{Partitioning for Noise Tolerance}

\begin{figure}
\centering
\includegraphics[width=3.1in]{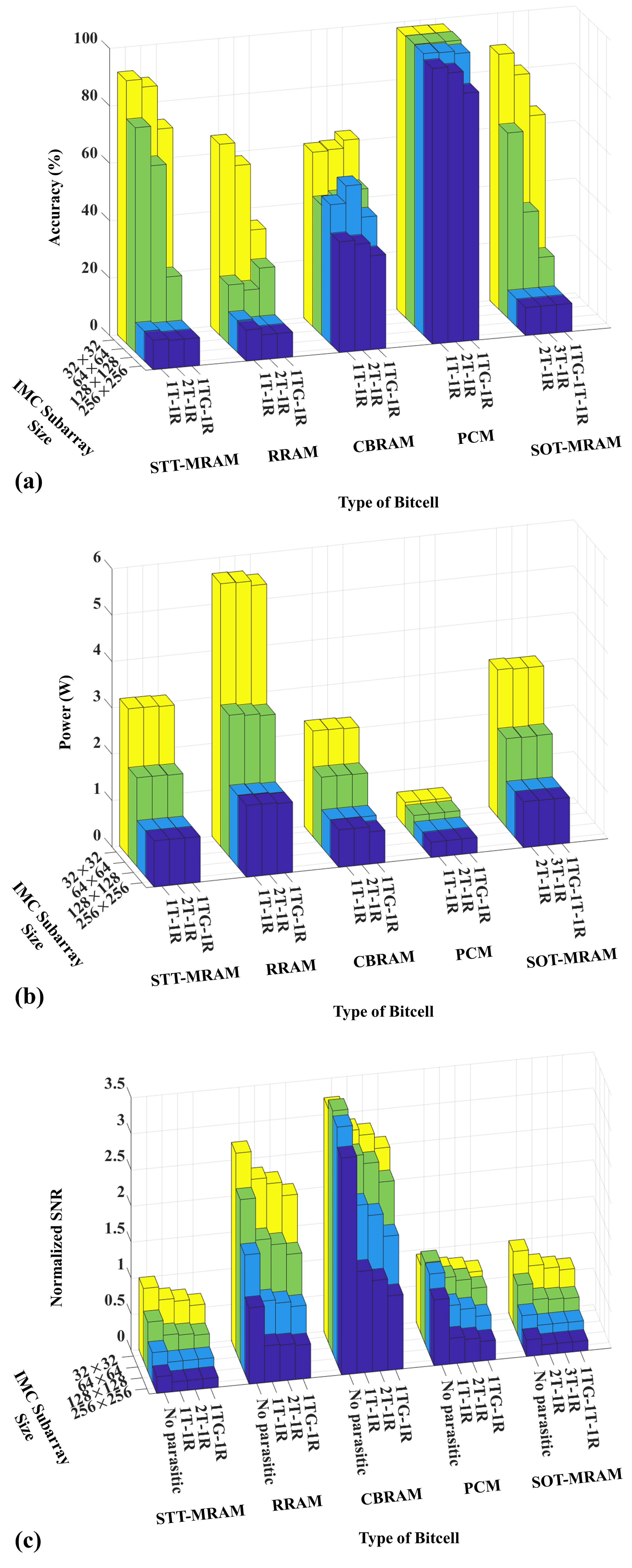}
\vspace{-3mm}
\caption{Results obtained for deploying 
the DNN on IMAC architectures with various subarray sizes, memristive technologies, and bitcell types. (a) Accuracy, (b) Power consumption, and (c) Normalized SNR values.} 
\label{fig:bar}
\end{figure}

For the noise tolerance analysis, we select the inputs of the differential amplifiers as one of the most sensitive nodes in the IMAC circuits. To measure the signal-to-noise (SNR) ratio in the IMAC circuits, we set all the inputs to VDD and all the weights to one ($G_{i,j}^+=1/R_{low}$ and $G_{i,j}^-=1/R_{high}$), and measure the average voltage difference between the output terminals of the $I^+$ and $I^-$ differential lines in IMC circuits and refer to it as \textit{Signal}. Figure \ref{fig:bar} (c) shows the SNR values that are all normalized with respect to the SNR of $32\times32$ subarray without parasitic. The results obtained show that Partitioning can increase the SNR and consequently noise tolerance for all cases with or without parasitics. This is because larger subarrays have larger number of resistive devices in the current paths which causes higher voltage drops and lower signal values at the output terminals of the crossbar. Moreover, it can be seen that memristive devices with lower $R_{high}/R_{low}$ ratio are more susceptible to noise. For instance, the highest SNR across all designs is achieved by $32\times 32$ IMC subarray with 1T-1R CBRAM bitcell \cite{shi2018neuroinspired} with $R_{high}/R_{low}$ ratio of 100, which is roughly 3.5$\times$ larger than the SNR realized by a similar design using MRAM technology with $R_{high}/R_{low}$=3. 

\vspace{-1mm}

\bibliographystyle{IEEEtran}

\balance
\bibliography{ref}
%



\end{document}